\documentclass{article} 
\usepackage{iclr2022_conference,times}

\usepackage{hyperref}
\usepackage{url}
\usepackage{tabulary}
\usepackage{multirow}
\usepackage{graphicx}
\usepackage{kotex}
\usepackage{xspace}
\usepackage{wrapfig}
\usepackage{amsmath,amssymb,amsfonts}

\title{Learning to Remember Patterns: Pattern Matching Memory Networks for Traffic Forecasting}

\author{Hyunwook Lee, Seungmin Jin, Hyeshin Chu, Hongkyu Lim, and Sungahn Ko\thanks{Corresponding Author}\\
Ulsan National Institute of Science and Technology\\
\texttt{\{gusdnr0916,skyjin,hyeshinchu,limhongkyu1219,sako\}@unist.ac.kr}\\
}

\newcommand{\rev}[1]{{#1}}

\newcommand{\toolname}{PM-MemNet\xspace}
\newcommand{\memory} {GCMem\xspace}
\iclrfinalcopy 
\begin{document}

\maketitle

\begin{abstract}

Traffic forecasting is a challenging problem due to complex road networks and sudden speed changes caused by various events on roads. Several models have been proposed to solve this challenging problem, with a focus on learning the spatio-temporal dependencies of roads. In this work, we propose a new perspective for converting the forecasting problem into a pattern-matching task, assuming that large traffic data can be represented by a set of patterns. To evaluate the validity of this new perspective, we design a novel traffic forecasting model called Pattern-Matching Memory Networks (\toolname), which learns to match input data to representative patterns with a key-value memory structure. We first extract and cluster representative traffic patterns that serve as keys in the memory. Then, by matching the extracted keys and inputs, \toolname acquires the necessary information on existing traffic patterns from the memory and uses it for forecasting. To model the spatio-temporal correlation of traffic, we proposed a novel memory architecture, \memory, which integrates attention and graph convolution. The experimental results indicate that \toolname is more accurate than state-of-the-art models, such as Graph WaveNet, with higher responsiveness. We also present a qualitative analysis describing how \toolname works and achieves higher accuracy when road speed changes rapidly.
\end{abstract}

\section{Introduction}

Traffic forecasting is a challenging problem due to complex road networks, varying patterns in the data, and intertwined dependencies among models. 
This implies that prediction methods should not only find intrinsic spatio-temporal dependencies among many roads, but also quickly respond to irregular congestion and various traffic patterns~\citep{Lee20} caused by external factors, such as accidents or weather conditions~\citep{Vlahogianni14, Li18, Xie20, Jiang21}. 
To resolve these challenges and successfully predict traffic conditions, many deep learning models have been proposed. 
Examples include the models with graph convolutional neural networks (GCNs)~\citep{Bruna14} and recurrent neural networks (RNNs)~\citep{Siegelmann91}, which outperform conventional statistical methods such as autoregressive integrated moving average (ARIMA)~\citep{Vlahogianni14, Li18dcrnn}.
Attention-based models, such as GMAN~\citep{Zheng20gman}, have also been explored to better handle complex spatio-temporal dependency of traffic data. 
Graph WaveNet~\citep{Wu19gwnet} adopts a diffusion process with a self-learning adjacency matrix and dilated convolutional neural networks (CNNs), achieving state-of-the-art performance. 
Although effective, existing models have a weakness in that they do not accurately forecast when conditions are abruptly changed (e.g., rush hours and accidents).

\begin{figure}[t]
    \centering
    \includegraphics[width=0.7\columnwidth]{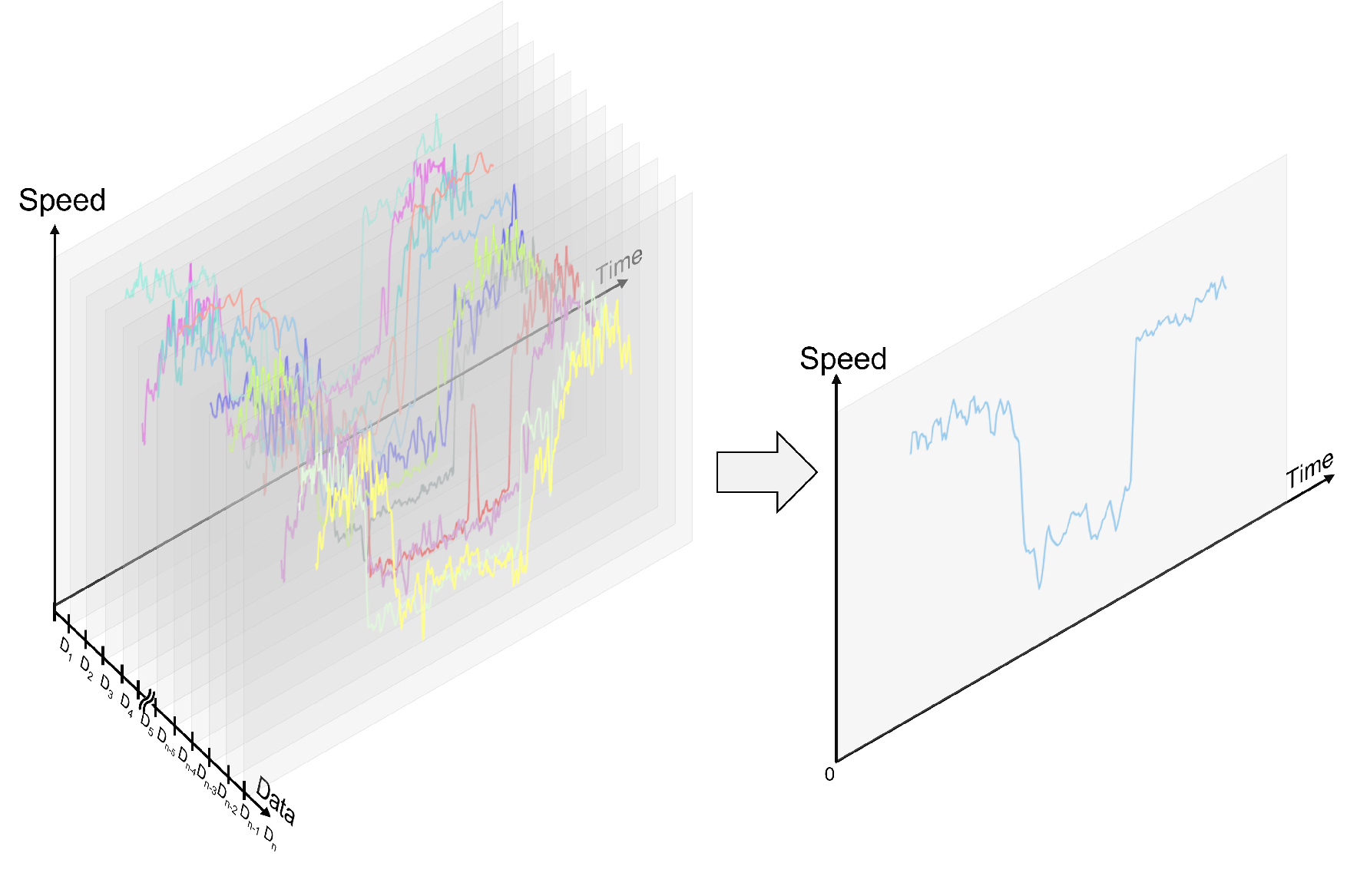}
    \caption{(left) Multiple traffic data with similar pattern, (right) extracted representative pattern}
    \label{fig:intro}
\end{figure}

In this work, we aim to design a novel method for modeling the spatio-temporal dependencies of roads and to improve forecasting performance. 
To achieve this goal, we first extract representative traffic patterns from historical traffic data, as we find that there are similar traffic patterns among roads, and a set of traffic patterns can be generalized for roads with similar spatio-temporal features. 
Figure~\ref{fig:intro} shows the example speed patterns (left, 90-minute window) that we extract from many different roads and a representative traffic pattern (right time series). 
With the representative patterns, we transform the conventional forecasting problem into a pattern-matching task to find out which pattern would be the best match for the given spatio-temporal features to predict future traffic conditions.
With insights from the huge success of neural memory networks in natural language processing and machine translation~\citep{Weston15, Sukhbaatar15, Kaiser17, Madotto18}, we design graph convolutional memory networks called \memory to manage representative patterns in spatio-temporal perspective.
Lastly, we design \toolname, which utilizes representative patterns from \memory for traffic forecasting. 
\toolname consists of an encoder and a decoder. 
The encoder consists of temporal embedding with stacked \memory, which generates meaningful representations via memorization, and the decoder is composed of a gated recurrent unit (GRU) with \memory.
We compare \toolname to existing state-of-the-art models and find that \toolname outperforms existing models. 
We also present a qualitative analysis in which we further investigate the strengths of \toolname in managing a traffic pattern where high responsiveness of a model to abrupt speed changes is desired for accurate forecasting.

The experimental results indicate that \toolname achieves state-of-the-art performance, especially in long-term prediction, compared to existing deep learning models. 
To further investigate the characteristics of \toolname, we conduct an ablation study with various decoder architectures and find that \toolname demonstrates the best performance. \rev{We also investigate how the number of representative patterns affects model performance.}
Finally, we discuss the limitations of this work and future directions for neural memory networks in the traffic forecasting domain.

The contributions of this work include: (1) computing representative traffic patterns of roads, (2) design of \memory to manage the representative patterns, (3) presenting a novel traffic prediction model, \toolname, that matches and uses the most appropriate patterns from \memory for traffic forecasting, (4) evaluation of \toolname compared to state-of-the-art models, (5) qualitative analysis to identify the strengths of \toolname, and (6) discussion of limitations and future research directions. 

\section{Related Work}

\subsection{Traffic Forecasting}
Deep learning models achieve huge success by effectively capturing spatio-temporal features in traffic forecasting tasks. Past studies ahve shown that RNN-based models outperform conventional temporal modeling approaches, such as ARIMA and support vector regression (SVR)~\citep{Vlahogianni14,Li18dcrnn}. 
More recently, many studies have demonstrated that attention-based models~\citep{Zheng20gman,Park20stgrat} and CNNs~\citep{Yu18stgcn,Wu19gwnet} record better performance in long-term period prediction tasks, compared to RNN-based models. 
In terms of spatial modeling,~\citet{Zhang16} propose a CNN-based spatial modeling method for Euclidean space. 
Another line of modeling methods, such as GCNs, using graph structures for managing complex road networks also become popular.
However, there are difficulties in using GCNs in the modeling process, such as the need to build an adjacency matrix and the dependence of GCNs on invariant connectivity in the adjacency matrix. 
To overcome these difficulties, a set of approaches, such as graph attention models (GATs), have been proposed to dynamically calculate edge importance~\citep{Park20stgrat}. 
GWNet~\citep{Wu19gwnet} adopts a self-adaptive adjacency matrix to capture hidden spatial dependencies in training. 
Although effective, forecasting models still suffer from inaccurate predictions due to abruptly changing road speeds and instability, with lagging patterns in long-term periods. 
To address these challenges, we build, save, and retrieve representative traffic patterns for predicting speed rather than directly forecasting with an input sequence.

\subsection{Neural Memory Networks}
Neural memory networks are widely used for sequence-to-sequence modeling in the natural language processing and machine translation domains. 
Memory networks are first proposed by \citet{Weston15} to answer a query more precisely even for large datasets with long-term memory. 
Memory networks perform read and write operations for given input queries. 
\citet{Sukhbaatar15} introduce end-to-end memory networks that can update memory in an end-to-end manner. Through the end-to-end memory learning, models can be easily applied to realistic settings. 
Furthermore, by using adjacent weight tying, they can achieve recurrent characteristics that can enhance generalization. 
\citet{Kaiser17} propose novel memory networks that can be utilized in various domains where life-long one-shot learning is needed. 
\citet{Madotto18} also introduce Mem2Seq, which integrates the multi-hop attention mechanism with  memory networks.
In our work, we utilize memory networks for traffic pattern modeling due to the similarity of the tasks and develop novel graph convolutional memory networks called \memory to better model the spatio-temporal correlation of the given traffic patterns.

\section{Proposed Approach}
In this section, we define the traffic forecasting problem, describe how we extract key patterns in the traffic data that serve as keys, and introduce our model, \toolname.

\subsection{Problem Setting}
To handle the spatial relationships of roads, we utilize a road network graph. 
We define a road network graph as $\mathcal{G} = (\mathcal{V}, \mathcal{E}, \mathcal{A})$, where $\mathcal{V}$ is a set of all different nodes with $|\mathcal{V}|=N$, $\mathcal{E}$ is a set of edges representing the connectivity between nodes, and $\mathcal{A}\in \mathbb{R}^{N\times N}$ is a weighted adjacency matrix that contains the connectivity and edge weight information. 
An edge weight is calculated based on the distance and direction of the edge between two connected nodes. 
As used in the previous approaches~\citep{Li18dcrnn,Wu19gwnet,Zheng20gman, Park20stgrat}, we calculate edge weights via the Gaussian kernel as follows: $A_{i,j} = \exp\big({-\frac{{\text{dist}}_{ij}^2}{\sigma^2}}\big)$, where dist$_{ij}$ is the distance between node $i$ and node $j$ and $\sigma$ is the standard deviation of the distances.

Prior research has formulated a traffic forecasting problem as a simple spatio-temporal data prediction problem~\citep{Li18dcrnn,Wu19gwnet,Zheng20gman,Park20stgrat} aiming to predict values in the next $T$ time steps using previous $T'$ historical traffic data and an adjacency matrix.
Traffic data at time $t$ is represented by a graph signal matrix, $X_\mathcal{G}^t \in \mathbb{R}^{N \times d_{in}}$, where $d_{in}$ is the number of features, such as speed, flow, and time of the day. 
In summary, the goal of the previous work is to learn a mapping function $f(\cdot)$ to directly predict future $T$ graph signals from $T'$ historical input graph signals:
\begin{align*}
    \big[{X_\mathcal{G}}^{(t-T'+1)}, \dots, {X_\mathcal{G}}^{(t)}\big] &\xrightarrow{f(\cdot)} \big[{X_\mathcal{G}}^{(t+1)}, \dots, {X_\mathcal{G}}^{(t+T)}\big]
\end{align*}

The goal of this study is different from previous work in that we aim to predict future traffic speeds from patterned data, instead of utilizing input $X_\mathcal{G}$ directly. 
\rev{We denote $\mathbb{P} \subset \mathbb{R}^{T'}$ as a set of representative traffic patterns, $p \in \mathbb{P}$ as one traffic pattern in $\mathbb{P}$, and $d: \mathbb{X} \times \mathbb{P} \rightarrow{[0,\infty)}$ as a distance function for pattern matching. 
Detailed information about traffic pattern extraction will be discussed in the next subsection.}
Our problem is to train the mapping function $f(\cdot)$ as follows:
\begin{align*}
    \big[{X_\mathcal{G}}^{(t-T'+1)}, \dots, {X_\mathcal{G}}^{(t)}\big] &\xrightarrow{d(\cdot), k-NN}
    \big[P^t_1, \dots, P^t_N\big] \xrightarrow{f(\cdot)}
    \big[{X_\mathcal{G}}^{(t+1)}, \dots, {X_\mathcal{G}}^{(t+T)}\big],
\end{align*}
where $P^t_i=\{p_1, \dots, p_k\}$ is \rev{a set of} $k$-nearest neighboring traffic patterns of node $i$ in time $t$, with a distance function $d$. Note that $p_j$ is the $j$-th nearest neighbor pattern. 

\begin{figure}[t]
    \centering
    \includegraphics[width=1\columnwidth]{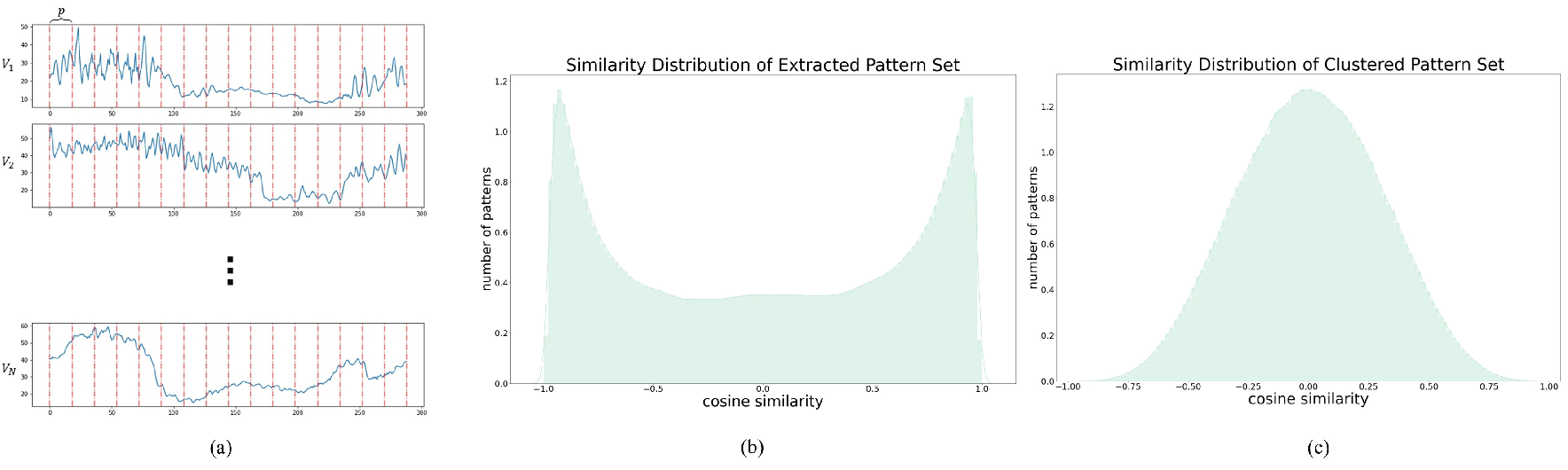}
    \caption{\rev{(a) Example daily patterns. Each part between the red dash lines is the speed patterns sliced by a given time window, (b) cosine similarity distribution of the original pattern set with class imbalance, and (c) the clustered pattern set.}}
    \label{fig:similarity measurement}
\end{figure}

\subsection{Key Extraction from Traffic Patterns}
\label{section:extraction}
\rev{Analyzing the traffic data, we find that the data has repeating patterns. 
In traffic data, the leading and trailing patterns have a high correlation, even during short-term periods. To take advantage of these findings, in our model, we build a representative pattern set, $\mathbb{P}$. 
First, from historical data, we compute an average daily pattern, which consists of 288 speed data points (total 24 hours with 5-minute intervals) for each vertex $v \in \mathcal{V}$. 
We then extract pattern $p$ by slicing the daily patterns with a given window size $T'$, as shown in Figure~\ref{fig:similarity measurement} (a). 
At this stage, $|\mathbb{P}| = N \times \lfloor \frac{288}{T'}\rfloor$. 
After we collect the patterns, we investigate similarity distribution of the extracted pattern set, $\mathbb{P}$, via cosine similarity (Figure~\ref{fig:similarity measurement} (b)) and find that the pattern set $\mathbb{P}$ has a biased distribution with too many similar patterns (i.e., class imbalance). 
Since such class imbalance causes memory ineffectiveness in accurate memory retrieval and gives biased training results, we use clustering-based undersampling~\citep{Lin17} with cosine similarity.
For example, if pattern $p$ and pattern $p'$ have a cosine similarity larger than $\delta$, they are in same cluster. We utilize the center of each cluster as a representative pattern of that cluster.
After undersampling by clustering, we have a balanced and representative pattern set, $\mathbb{P}$, as shown in Figure~\ref{fig:similarity measurement} (c), which we use as keys for memory access. 
Table~\ref{tab:ablation} presents the effect of different $\delta$ and $|\mathbb{P}|$ on forecasting performance.}

\subsection{Neural Memory Architecture}
\begin{figure}[t]
    \centering
    \includegraphics[width=1\columnwidth]{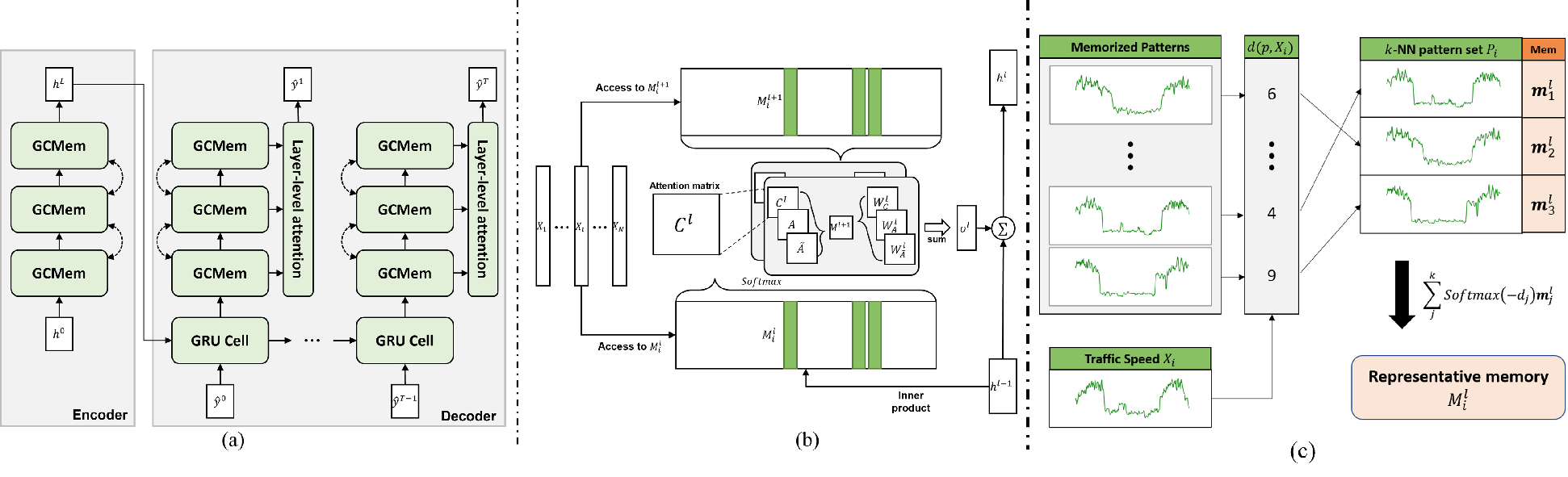}
    \caption{\rev{(a) Overall architecture of \toolname where $L = 3$. Dashed line means adjacent weight tying. (b) \memory architecture with GCNs (gray blocks) and (c) representative memory selection among $k$-nearest patterns for $X_i$ with $k$=3, $d_j = d(p_j,X_i)$.}}
    \label{fig:model}
\end{figure}

Conventionally, memory networks have used the attention mechanism for memory units to enhance memory reference performance, but this attention-only approach cannot effectively capture spatial dependencies among roads.
To address this issue, we design a new memory architecture, \memory (Figure~\ref{fig:model} (b)), which integrates multi-layer memory with the attention mechanism~\citep{Madotto18} and graph convolution~\citep{Bruna14}. 
By using \memory, a model can capture both pattern-level attention and graph-aware information sharing via GCNs.

To effectively handle representative patterns from a spatio-temporal perspective, we utilize several techniques. 
\rev{First, we use a modified version of the adjacent weight tying technique in MemNN~\citep{Sukhbaatar15, Madotto18}, which has been widely used for sentence memorization and connection searches between query and memorized sentences.}
\citet{Sukhbaatar15,Madotto18} propose the technique to capture information from memorized sentences by making use of sentence-level attention. 
However, their methodology only learns pattern similarity and cannot handle spatial dependency. 
Using the same method in traffic forecasting is insufficient since handling a graph structure is essential for building spatial dependencies of roads. 
In order to consider the graph structure while maintaining the original sentence-level attention score, we use an adjacency matrix, a learnable adaptive matrix, and attention scores for the GCNs. 
\rev{By maintaining pattern-level attention, the model takes advantage of both pattern-level information sharing and adjacent weight tying~\citep{Madotto18}.}
As a result, adjacent memory cells can effectively retain attention mechanisms while considering a graph structure. 

Figure~\ref{fig:model} (a) and (b) shows \toolname and our proposed graph convolution memory architecture.
\rev{The memories for the \memory are embedding matrices $\mathbf{M} = \{\mathbf{M}^1,\dots,\mathbf{M}^{L+1}\}$, where $\mathbf{M}^l \in \mathbb{R}^{|\mathbb{P}| \times d_h}$. }
For memory reference, we utilize pattern set $\mathbb{P}$, which contains the extracted traffic patterns (in Section~\ref{section:extraction}) and $k$-Nearest Neighbor ($k$-NN) with distance function, $d(\cdot)$. 
For each input traffic data $X_i \in \mathbb{R}^{T' \times d_{in}}$ in node $i$, based on $k$-nearest patterns and the distance function, we build a representative memory $M^l_i$ as follows:
\rev{
\begin{align}
    M^l_i &= \sum^k_{j=1} \text{Softmax}(-d_j)\mathbf{m}^l_j,
\end{align}}
where \rev{$\mathbf{m}^l_j = \mathbf{M}^l(p_j)$} is the memory context for $p_j$ and layer $l$, $d_j = g(X_i, p_j)$, and Softmax$(z_i)=e^{z_i} / \sum_j e^{z_j}$. We summarize our representative memory selection process in Figure~\ref{fig:model} (c).

\rev{After calculating representative memory, \memory calculates hidden state $h^l \in \mathbb{R}^{N \times d_h}$ with previous hidden state $h^{l-1}$ and representative memory $M^l$ and $M^{l+1}$ (Figure~\ref{fig:model} (b)).}
For each representative memory $M^l_j$, \toolname calculates pattern-level attention scores, $ \alpha^l_{i,j}$ as follows:
\begin{align}
    \alpha^l_{i,j} &= \text{Softmax}(\frac{h_{i}^{l-1} (M^l_j)^\top}{\sqrt{d_{h}}}),\label{eq:attention}
\end{align}
where $h_{i}^{l-1} \in \mathbb{R}^{d_h}$ is the previous hidden state of node $i$. We denote an attention matrix as $C^l \in \mathbb{R}^{N \times N}$, where \rev{$\alpha^l_{i,j}$ is the entry in the $i$-th row and $j$-th column of $C^l$.} Then, given memory unit $M^{l+1} \in \mathbb{R}^{N \times d_h}$ for the next layer $l+1$, we calculate output feature, $o^{l}$ with graph convolution as shown below:
\begin{align}
    o^l &= \sum_i^{h}\big(W^l_{A,i}M^{l+1}A + W^l_{\tilde{A},i}M^{l+1}\tilde{A} + W^l_{C,i}M^{l+1}C^l\big),
\end{align}
where $A \in \mathbb{R}^{N \times N}$ is an adjacency matrix and $\tilde{A} = \text{Softmax}(ReLU(E_1E_2^\top))$ is a learnable matrix, which captures hidden spatio-temporal connections~\citep{Wu19gwnet, Shi19}. 
\rev{$E_1, E_2 \in \mathbb{R}^{N \times d_h}$ are learnable node embedding vectors and $W \in \mathbb{R}^{d_h}$ is a learnable matrix.}
We use $ReLU$ as an activation function. 
Then, we update hidden states by $h^l = h^{l-1} + o^l$. 
Before we update the hidden states, we apply batch normalization on $o^l$.

\subsection{Encoder Architecture}
\rev{The left side of Figure~\ref{fig:model} (a) shows the proposed encoder architecture.}
\toolname handles traffic patterns and their corresponding memories. 
Although the patterns in the memory provide enough information for training, there are other types of data that can also be used for prediction. 
For example, different roads have their own patterns that may not be captured in advance (e.g., the unique periodicity of roads around an industry complex~\citep{Lee20}). 
In addition, there would be some noise and anomalous patterns due to various events (e.g., accidents), which are not encoded when patterns are grouped. 
As such, we provide embedding for the time (i.e., $emb$) and noise (i.e., $N_i$) that the encoder uses to generate input query $h^0$ for \memory. 
Specifically, for the time series $\mathbf{T} = [t-T'+1, \dots, t]$ and noise $N_i = X_i - p_{1}$, we calculate its representation, $h^0_i$ as shown below:
\begin{align}
    h^0_i &= emb(T) + W_n N_i,
\end{align}
where $emb$ and $W_n$ represent a learnable embedding for the time of day and a learnable matrix, respectively. 
\rev{Note that $emb(T) \in \mathbb{R}^{d_h}$ and $W_n \in \mathbb{R}^{d_h \times T}$}
\toolname updates $h^0_i$ using $L$-layer \memory. 
\rev{We use the output of the encoder $h^L \in \mathbb{R}^{N \times d_h}$ as an initial hidden state in the decoder.}

\subsection{Decoder Architecture}
As shown in Figure~\ref{fig:model} (a), we build our decoder architecture using single layer GRU, followed by the $L$-stacked \memory. 
\rev{For each $t$ step, the decoder predicts the value at time step $t$ using the previous prediction $\hat{y}^{t-1}$ and GRU hidden state $\tilde{h}_{t-1}$. Initially, $\hat{y}^{0}$ is zero matrix and $\tilde{h}_0$ is encoder hidden state $h^L$.}
The hidden states from the GRU will be an input for the $L$-stacked \memory. 
Similar to our encoder architecture, \memory updates hidden states with attention and GCNs. 
Instead of using updated hidden states for prediction directly, we utilize layer-level self-attention in the decoder. Specifically, for each \memory layer $l$ and node $i$, we calculate energy, $e_{i,l}$ using the previous hidden state $h^{l-1}_i$ and memory context $M^l_i$ as shown below:
\begin{align}
    e_{i,l} = \frac{(h^{l-1}_i) (M^l_i W^l)^\top}{\sqrt{d_h}},
\end{align}
where $d_h$ is the hidden size and $W^l \in mathbb{R}^{d_h \times d_h}$ is learnable matrix. Then, with the output feature $o^l_i$ of each layer $l$ and node $i$, we can predict $\hat{y}^t_i$ as:
\begin{align}
    \hat{y}_i^t &= \sum_l^L \alpha_{i,l} o^l_i W_{proj},
\end{align}
where $W_{proj} \in \mathbb{R}^{d_h \times d_{out}}$ is a projection layer and $\alpha_{i,l} = \text{Softmax}(e_{i,l})$.
\rev{Note that $h^0$ is equivalent to a hidden state of a GRU cell, $\tilde{h}_t$.}
Using layer-level attention, \toolname utilizes information from each layer more effectively.

\section{Evaluation}
\label{section:evaluation}
In this section, we explain the experiments conducted to compare \toolname to existing models in terms of accuracy. 
We use two datasets for the experiment--METR-LA and NAVER-Seoul\textsuperscript{\ref{github}}.
METR-LA contains 4-month speed data from 207 sensors of Los Angeles highways~\citep{Li18dcrnn}. 
NAVER-Seoul has 3-month speed data collected from 774 links in Seoul, Korea. 
As NAVER-Seoul data covered the main arterial roads in Seoul, it can be considered a more difficult dataset with many abruptly changing speed patterns compared to METR-LA data. 
Both datasets have five-minute interval speeds and timestamps. 
Before training the \toolname, we have filled out missing values using historical data and applied z-score normalization. 
We use 70\% of the data for training, 10\% for validation, and the rest for evaluation, as \citet{Li18dcrnn, Wu19gwnet, Zheng20gman} have done in their work.

\subsection{Experimental Setup}
In our experiment, we use 18 sequence data points as a model input ($T' = 18$, one and a half hours) and predict the next 18 sequences. 
For the $k$-NN, we utilize the cosine similarity function for the similarity measurement and set $k = 3$. 
To extract $\mathbb{P}$, we utilize the training dataset and initialize parameters and embedding using Xavier initialization.
After performing a greedy search among $d_h = [16, 32, 64, 128]$, $L = [1, 2, 3, 4]$, and \rev{various $|\mathbb{P}|$ values, we set $d_h$ as 128, $L$ as 3, and $|\mathbb{P}| \approx 100$ with $\delta = 0.7$ and $0.9$ for NAVER-Seoul and METR-LA, respectively. 
Table~\ref{tab:ablation} presents the experimental results with different memory and pattern sizes (i.e., $|\mathbb{P}|$)}
We apply the Adam optimizer with a learning rate of 0.001 and use the mean absolute error (MAE) as a loss function.

\rev{We compare \toolname to the following baseline models: (1) multilayer perceptron (MLP); (2) STGCN~\citep{Yu18stgcn}, which forecasts one future step using graph convolution and CNNs; (3) Graph Convolution Recurrent Neural Network (GCRNN); (4) DCRNN~\citep{Li18dcrnn}, a sequence-to-sequence model that combines diffusion convolution in the gated recurrent unit; (5) ASTGCN~\citep{Guo19astgcn}, which integrates GCN, CNN, and spatial  and temporal attention; (6) Graph-WaveNet (GWNet)~\citep{Wu19gwnet}, which forecasts multiple steps at once by integrating graph convolution and dilated convolution; and (7) Graph Multi-Attention Network (GMAN)~\citep{Zheng20gman} which integrates spatial and temporal attention with gated fusion mechanism. 
GMAN also predicts multiple steps at once. }
To allow detailed comparisons, we train the baseline models to forecast the next 90 minutes of speeds at 5-minute intervals, given the past 18 5-minute interval speed data points.
We train the baseline models in an equivalent environment with PyTorch\footnote{\url{https://pytorch.org}} using the public source codes and settings provided by the authors. 
\rev{Detailed settings, including the hyperparameters, are available in Appendix~\ref{section:detailed_setting}.}
\rev{To ease the performance comparison with previous work, we provide additional experimental results with a commonly used setting, where $T' = T = 12$, in Table~\ref{table:common_setting} in Appendix.}
\rev{We also present a qualitative analysis of our method for NAVER-Seoul dataset in Appendix~\ref{section:qualitative}}
Our source codes are available on GitHub\footnote{\label{github} \url{https://github.com/HyunWookL/PM-MemNet}}.

\subsection{Experimental Results}
\begin{table*}[t]
\centering
\caption{Experimental Results for NAVER-Seoul and METR-LA datasets}
\resizebox{\linewidth}{!}{\begin{tabular}{c|c|c|c|c|c|c|c|c|c|c|c|}
\hline
Dataset                & T                       & Metric & HA         & MLP & STGCN & GCRNN     & DCRNN & ASTGCN & GWNet & GMAN & \toolname \\
\hline\hline 
 \multirow{12}{*}{NAVER-Seoul}    & \multirow{3}{*}{15min}  & MAE    &  6.54 &  5.28 &  4.63 &  4.87 &  4.86 &  5.09 &   4.91          &  5.20 &  \textbf{4.57}\\
                                  &                         & MAPE   & 18.24 & 16.86 & 14.49 & 15.23 & 15.35 & 16.14 &  14.86          & 16.98 & \textbf{14.43}\\
                                  &                         & RMSE   &  9.32 &  7.78 &  6.92 &  7.18 &  7.12 &  7.44 &   7.24          &  8.32 &  \textbf{6.72}\\
                                  \cline{2-12}
                                  & \multirow{3}{*}{30min}  & MAE    &  7.16 &  6.13 &  5.50 &  5.73 &  5.67 &  5.71 &   5.26          &  5.35 &  \textbf{5.04}\\
                                  &                         & MAPE   & 20.15 & 20.05 & 17.37 & 18.17 & 18.38 & 18.78 &  \textbf{16.16} & 17.47 & 16.34\\
                                  &                         & RMSE   & 10.18 &  9.51 &  8.83 &  9.03 &  8.80 &  8.73 &   8.13          &  8.67 &  \textbf{7.86}\\
                                  \cline{2-12}
                                  & \multirow{3}{*}{60min}  & MAE    &  8.22 &  7.08 &  6.77 &  6.58 &  6.40 &  6.22 &   5.55          &  5.48 &  \textbf{5.24}\\
                                  &                         & MAPE   & 23.37 & 23.44 & 20.42 & 20.95 & 21.09 & 20.37 &  16.97          & 17.89 & \textbf{16.94}\\
                                  &                         & RMSE   & 11.54 & 11.13 & 10.89 & 10.58 & 10.06 &  9.58 &   8.77          &  8.94 &  \textbf{8.39}\\
                                  \cline{2-12}
                                  & \multirow{3}{*}{90min}  & MAE    &  9.24 &  7.79 &  8.06 &  7.14 &  6.86 &  6.76 &   5.87          &  5.58 &  \textbf{5.40}\\
                                  &                         & MAPE   & 26.40 & 26.08 & 22.93 & 22.86 & 22.74 & 21.83 &  17.89          & 18.18 & \textbf{17.44}\\
                                  &                         & RMSE   & 12.77 & 12.17 & 12.86 & 11.43 & 10.69 & 10.32 &   9.33          &  9.09 &  \textbf{8.68}\\
                                  \cline{1-12}

\multirow{12}{*}{METR-LA}    & \multirow{3}{*}{15min}   & MAE    &  4.23 &  2.93 &  2.61 &  2.59 &  \textbf{2.56} &  3.25 &  2.72 &  2.86 &  2.66 \\
                             &                         & MAPE    &  9.76 &  7.76 &  \textbf{6.59} &  6.73 &  6.67 &  9.27 &  7.14 &  7.67 &  7.06 \\
                             &                         & RMSE    &  7.46 &  5.81 &  5.19 &  \textbf{5.12} &  5.10 &  6.28 &  5.20 &  5.77 &  5.28 \\
                             \cline{2-12}
                             & \multirow{3}{*}{30min}  & MAE     &  4.80 &  3.60 &  3.22 &  3.08 &  \textbf{3.01} &  3.80 &  3.12 &  3.14 &  3.02 \\
                             &                         & MAPE    & 11.30 & 10.00 &  \textbf{8.39} &  8.72 &  8.42 & 11.28 &  8.66 &  8.79 &  8.49 \\
                             &                         & RMSE    &  8.34 &  7.29 &  6.63 &  6.32 &  6.29 &  7.59 &  6.34 &  6.54 &  \textbf{6.28} \\
                             \cline{2-12}
                             & \multirow{3}{*}{60min}   & MAE    &  5.80 &  4.69 &  4.31 &  3.74 &  3.60 &  4.49 &  3.58 &  3.48 &  \textbf{3.40} \\
                             &                          & MAPE   & 14.04 & 13.68 & 11.13 & 11.50 & 10.73 & 13.69 & 10.30 & 10.10 &  \textbf{9.88} \\
                             &                          & RMSE   &  9.86 &  9.24 &  8.71 &  7.71 &  7.65 &  8.94 &  7.53 &  7.30 &  \textbf{7.24} \\
                             \cline{2-12}
                             & \multirow{3}{*}{90min}   & MAE    &  6.65 &  5.58 &  5.41 &  4.23 & 4.06  &  4.97 &  3.85 &  3.71 &  \textbf{3.64} \\
                             &                          & MAPE   & 16.37 & 17.08 & 13.76 & 13.49 &12.53  & 15.53 & 11.39 & 11.00 & \textbf{10.74} \\
                             &                          & RMSE   & 10.97 & 10.52 & 10.47 &  8.79 & 8.58  &  9.71 &  8.12 &  7.71 &  \textbf{7.74} \\
                             
\hline
\end{tabular}}
\label{table:model_evaluation}
\end{table*}

Table~\ref{table:model_evaluation} displays the experimental results for NAVER-Seoul and METR-LA for the next 15 minutes, 30 minutes, 60 minutes, and 90 minutes using mean absolute error (MAE), mean absolute percentage error (MAPE), and root mean square error (RMSE). 
\rev{Note that Table~\ref{table:detailed_exp} and Figure~\ref{fig:errorbar} in Appendix show more detailed experimental results and error information. }
As Table~\ref{table:model_evaluation} shows, \toolname achieves state-of-the-art performance for both datasets. Specifically, \toolname yields the best performance in all intervals with NAVER-Seoul data, while it outperforms other models in the long-term prediction (i.e., 60 and 90 minutes) with METR-LA data.

In both datasets, we find interesting observations. 
First, RNN-based models perform better than other models for short term periods (i.e., 15 minutes), but they show a weakness for long-term periods. 
This problem occurs in sequence-to-sequence RNNs due to error accumulation caused by its auto-regressive property. 
Compared to the RNN-based models, \toolname shows a lesser performance decrease, although it had decoder based on RNN architecture.
This is because of \memory and pattern data, which make representations for traffic forecasting more robust than other methods. 
In the case of NAVER-Seoul, we observe that all models suffer from decreased accuracy due to more complicated urban traffic patterns and road networks than those in METR-LA.
In spite of these difficulties, \toolname proves its efficiency with representative traffic patterns and the memorization technique. This result strengthens our hypothesis that the traffic data used for traffic forecasting can be generalized with a small number of pattern even when the traffic data are complex and hard to forecast.

\begin{table*}[t]
    \centering
    \caption{Ablation study result. Note that `Ours' means \toolname.}
\resizebox{\linewidth}{!}{\begin{tabular}{c|c|c|c|c|c|c|c|c|c|}
\hline
Dataset                & T                       & Metric & Ours & SimpleMem & CNN Decoder & RNN Decoder & Ours (L=1)  & Ours ($|\mathbb{P}| = k$) & Ours ($|\mathbb{P}| >> 1000$)\\
\hline\hline 
\multirow{12}{*}{NAVER-Seoul}     & \multirow{3}{*}{15min}  & MAE    &  4.57 &  5.72 &  \textbf{4.56} &  4.67 &  4.72           &  4.66 &   4.59\\
                                  &                         & MAPE   & 14.43 & 18.18 & \textbf{14.40} & 14.84 & 14.98           & 14.77 &  14.48\\
                                  &                         & RMSE   &  6.72 &  8.79 &  \textbf{6.71} &  6.83 &  6.87           &  6.89 &   6.73\\
                                  \cline{2-10}
                                  & \multirow{3}{*}{30min}  & MAE    &  \textbf{5.04} &  5.84 &  5.06 &  5.19 &  5.22           &  5.21 &   5.09\\
                                  &                         & MAPE   & \textbf{16.34} & 18.86 & 16.36 & 16.87 & 16.97           & 16.91 &  16.41\\
                                  &                         & RMSE   &  \textbf{7.86} &  9.24 &  7.90 &  8.02 &  8.03           &  8.18 &   7.97\\
                                  \cline{2-10}
                                  & \multirow{3}{*}{60min}  & MAE    &  \textbf{5.24} &  6.38 &  5.32 &  5.47 &  5.52           &  5.53 &   5.36\\
                                  &                         & MAPE   & \textbf{16.94} & 21.42 & 17.19 & 17.91 & 17.97           & 18.05 &  17.19\\
                                  &                         & RMSE   &  \textbf{8.39} & 10.08 &  8.51 &  8.69 &  8.70           &  8.92 &   8.67\\
                                  \cline{2-10}
                                  & \multirow{3}{*}{90min}  & MAE    &  \textbf{5.40} &  6.95 &  5.55 &  5.70 &  5.72           &  5.74 &   5.52\\
                                  &                         & MAPE   & \textbf{17.44} & 23.89 & 17.99 & 18.73 & 18.63           & 18.73 &  17.76\\
                                  &                         & RMSE   &  \textbf{8.68} & 10.88 &  8.82 &  9.10 &  9.05           &  9.31 &   8.94\\
                                  \cline{2-10}
\cline{1-1}
 \multirow{12}{*}{METR-LA}        & \multirow{3}{*}{15min}  & MAE    &  2.66 &  3.01 &  \textbf{2.63} &  2.68 &  2.68               &  2.67 &  2.68\\ 
                                  &                         & MAPE   &  7.06 &  8.03 &  \textbf{6.98} &  7.10 &  7.11               &  7.09 &  7.13\\ 
                                  &                         & RMSE   &  \textbf{5.28} &  5.94 &  5.32 &  5.31 &  5.31               &  5.35 &  5.31\\ 
                                  \cline{2-10}
                                  & \multirow{3}{*}{30min}  & MAE    &  3.02 &  3.27 &  \textbf{3.01} &  3.06 &  3.06               &  3.06 &  3.04\\ 
                                  &                         & MAPE   &  8.49 &  9.20 &  \textbf{8.46} &  8.56 &  8.59               &  8.67 &  8.51\\ 
                                  &                         & RMSE   &  6.28 &  6.68 &  6.36 &  6.32 &  \textbf{6.27}               &  6.36 &  6.32\\ 
                                  \cline{2-10}
                                  & \multirow{3}{*}{60min}  & MAE    &  \textbf{3.40} &  3.72 &  3.41 &  3.46 &  3.47               &  3.49 &  3.45\\ 
                                  &                         & MAPE   &  \textbf{9.88} & 10.94 &  \textbf{9.88} & 10.02 & 10.07      & 10.34 &  9.87\\
                                  &                         & RMSE   &  \textbf{7.24} &  7.70 &  7.28 &  7.31 & 7.25                &  7.39 &  7.32\\ 
                                 \cline{2-10}
                                  & \multirow{3}{*}{90min}  & MAE    &  \textbf{3.64} &  4.09 &  3.65 &  3.71 &  3.73               &  3.75 &  3.69\\ 
                                  &                         & MAPE   & \textbf{10.74} & 12.25 & 10.85 & 10.87 & 10.98               & 11.30 & 10.63\\
                                  &                         & RMSE   &  7.74 &  8.38 &  \textbf{7.71} &  7.81 &  7.75               &  7.91 &  7.81\\ 
                                  \cline{2-10}

\hline
\end{tabular}}
\label{tab:ablation}
\end{table*}

\subsection{Ablation Study}
\label{section:ablation}

\begin{wrapfigure}{r}{0.42\textwidth}
  \begin{center}
    \includegraphics[width=0.40\textwidth]{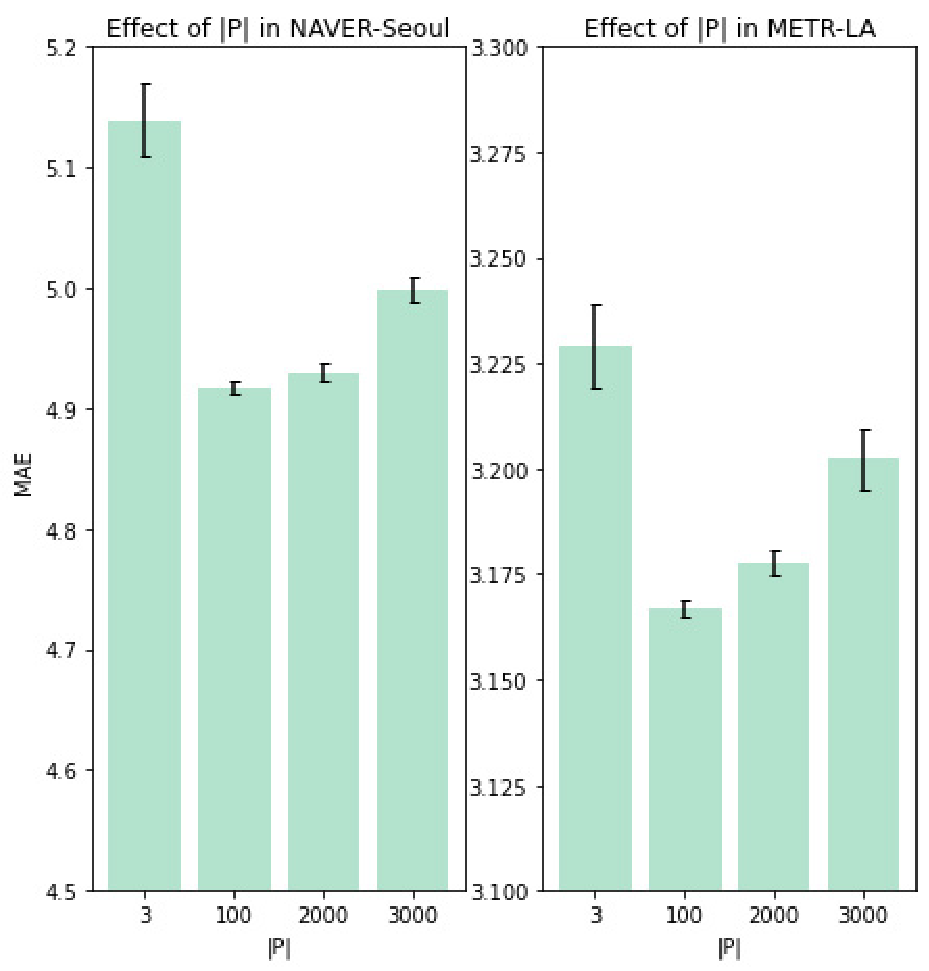}
  \end{center}
  \caption{Effect of $|\mathbb{P}|$ in NAVER-Seoul (left) and METR-LA (right).}
  \label{fig:ablation}
\end{wrapfigure}

We further evaluate our approach by conducting ablation studies (Table~\ref{tab:ablation}, Table~\ref{table:detailed_ablation} in Appendix).
First, we check whether \memory can effectively model spatio-temporal relationships among road networks. 
To this end, we compare the performance of \toolname to that of SimpleMem, a simplified version of \toolname in which the memory layer depends only on pattern-level attention and does not consider any graph-based relationship. 
By comparing \toolname and SimpleMem, we observe that SimpleMem shows nearly 10\% decrease in performance, which shows the importance of a graph structure in modeling traffic data. 
In the same context, according to Table~\ref{table:model_evaluation} and Table~\ref{tab:ablation}, SimpleMem has lower accuracy than previous models, such as DCRNN, which considers a graph structure. 

Next, we combine \memory with various decoder architectures to investigate how the performance of \memory changes with different decoder combinations.
The results indicate that all decoder combinations achieve state-of-the-art performance for both NAVER-Seoul and METR-LA datasets. 
This result implies that \memory effectively learns representations, even with a simple decoder architecture, such as single layer RNN. 
Also, from the perspective of error accumulation, we notice that \toolname outperforms the CNN decoder even for the 90-minute prediction. 
In contrast, the one-layered RNN decoder shows a lower performance than the CNN decoder. 
However, when comparing the RNN decoder to existing RNN-based models, RNN decoder can predict more precisely for the long-term prediction. 
These results also show that \memory is a robust solution for long-term dependency modeling.
Modifying the \memory layer depth, we discover that \toolname generates sufficiently accurate predictions with a single memory layer (i.e., \toolname w/L=1). 
Although \toolname needs a three-layered \memory to achieve the highest accuracy, we can still consider deploying the model with a lightweight version while ensuring its state-of-the-art performance. 

\rev{Next we analyze how different memory sizes (i.e., different pattern numbers) affect the performance. 
In this experiment, we set $|\mathbb{P}|$ as 3, 100, 2000, and 3000. 
Note that we set $|\mathbb{P}|$ as 3 to simulate an extreme case, where \toolname has very rare memory space.
Figure~\ref{fig:ablation} shows the experimental results. We can observe that \toolname records the best performance with 100 traffic patterns; attaching large $|\mathbb{P}|$ to a model does not always allow the best forecasting performance. 
We provide the time consumption of the models in Table~\ref{table:time} in Appendix.
}

\section{Limitations, Discussion, and Future Directions}
This work is the first attempt to design neural memory networks for traffic forecasting. 
Although effective, as shown in the evaluation results, there are limitations to this approach. 
First, we extract traffic patterns to memorize them in advance; however, it is possible that the extracted patterns are redundant, even after strict filtering. 
As such, there is need to find important patterns and optimize the number of memory slots. 
For example, a future study could investigate how to learn and extend the key space during the training phase~\citep{Kaiser17}.
Second, the learning of embedding matrices in \toolname proceeds only based on referred patterns. Because there are no further losses to optimize memory itself, patterns not referred to are not trained. This training imbalance among memories is of interest, as the model cannot generate meaningful representations from rare patterns. A future study may research not only how such a representation imbalance affects performance, but also design a loss function to reduce the representation gap between rare and frequent events.
Third, we use cosine similarity in this work, but it may not be an optimal solution, since it causes mismatching with noisy traffic data. Also, the optimal window size for pattern matching remains to be addressed.
A future study may focus on approaches to effectively compute the similarity of traffic patterns. 
Designing a learnable function for the distance measurement is one possible direction.
Finally, we show that the model can effectively forecast traffic data with a small group of patterns. 
This implies a new research direction for comparing results and learning methods that work with sparse data, such as meta learning and few or zero shot learning~\citep{Kaiser17}. 

\section{Conclusion}
In this work, we propose \toolname, a novel traffic forecasting model with a graph convolutional memory architecture, called \memory. 
By integrating GCNs and neural memory architectures, \toolname effectively captures both spatial and temporal dependency. 
By extracting and computing representative traffic patterns, we reduce the data space to a small group of patterns. 
The experimental results for METR-LA and NAVER-Seoul indicate that \toolname outperforms state-of-the-art models.
It proves our hypothesis ``accurate traffic forecasting can be achieved with a small set of representative patterns'' is reasonable. 
We also demonstrate that \toolname quickly responds to abruptly changing traffic patterns and achieves higher accuracy compared to the other models. 
Lastly, we discuss the limitations of this work and future research directions with the application of neural memory networks and a small set of patterns for traffic forecasting. 
In future work, we plan to conduct further experiments using \toolname for different spatio-temporal domains and datasets to investigate whether the insights and lessons from this work can be generalized to other domains.

\subsubsection*{Acknowledgements}
This work was supported by the Korean National Research Foundation (NRF) grant (No. 2021R1A2C1004542) and by the Institute of Information \& Communications Technology Planning \& Evaluation (IITP) grants (No. 2020-0-01336--Artificial Intelligence Graduate School Program (UNIST), No. 2021-0-01198--ICT R\&D Innovation Voucher Program), funded by the Korea government (MSIT). This work was also partly supported by NAVER Corp.

\bibliography{ms}
\bibliographystyle{iclr2022_conference}

\appendix
\section{Appendix}

\subsection{Detailed Experimental Setup}
\label{section:detailed_setting}

\rev{\textbf{MLP} Multi-layer perceptron with two hidden layers; each layer contains 64 units and rectified linear unit (ReLU) activation.}

\rev{\textbf{STGCN} Spatio-Temporal Graph Convolutional Networks~\citep{Yu18stgcn}. STGCN models spatial features using spectral-based graph convolution with Chebyshev polynomial approximation and temporal features using gated CNNs. In our experiment, like the original paper, we set STGCN with a graph convolution kernel size $K = 3$, temporal convolution kernel size of $K_t = 3$, and three layers with  64, 16, and 64 hidden units, respectively.}

\rev{\textbf{DCRNN} Diffusion Convolutional Recurrent Neural Networks~\citep{Li18dcrnn}. DCRNN is the model that integrates diffusion convolution with RNNs, especially the GRU. In our experiment, both encoder and decoder contain two recurrent layers with 64 hidden units and maximum steps of random walks set as 3 (i.e., $K = 3$).}

\rev{\textbf{GCRNN} Graph Convolutional Recurrent Neural Networks. GCRNN is a variant of DCRNN that integrates simple bidirectional graph convolution and sequence-to-sequence architecture. Both encoder and decoder contain two recurrent layers with 64 hidden units and maximum steps of random walks set as 3 (i.e., $K = 3$).}

\rev{\textbf{ASTGCN} Attention based Spatio-Temporal Graph Convolutional Networks~\citep{Guo19astgcn}. ASTGCN models spatio-temporal features by attention, GCNs, and CNNs. For the GCNs, ASTGCN utilizes spectral-based graph convolution with Chebyshev polynomials $K = 3$ and kernel size of CNNs $K_t = 3$. Furthermore, all layers have the same hidden unit size of 64.}

\rev{\textbf{GWNet} Graph WaveNet~\citep{Wu19gwnet}. GWNet combines graph convolution and dilated causal convolution. In our experiment, to cover the input sequence length, we utilize 12 layers with dilation factors of $1,2,1,2,\dots,1,2$. Also, following original paper, we set the diffusion step $K = 3$ and $E_1, E_2 \in \mathbb{R}^{N \times 10}$.}

\rev{\textbf{GMAN} Graph Multi-Attention Network~\citep{Zheng20gman}. GMAN handles both spatial and temporal features using attention and pretrained node embedding. In our experiment, we utilize the same setting as the original paper, that is, three layers for both encoder and decoder with eight attention heads and 64 hidden units (8 hidden units per attention head).}

\rev{\textbf{PM-MemNet} Pattern Matching Memory Networks. Both encoder and decoder have three GCMem layers with 128 hidden units and $|\mathbb{P}| \approx 100$. Note that the decoder also has one GRU cell before GCMem with 128 hidden units. For the $k$-NN, we utilize $k = 3$ and cosine similarity. Also, for the graph convolution, we set the diffusion step as $K = 2$.}

\rev{In the experiment, we utilize the MAE loss function and Adam optimizer with a learning rate of $1e^{-3}$ for training. The learning rate reduces to $\frac{1}{10}$ for every 10 epochs, starting at 30 epochs, until it reaches $1e^{-6}$. For each model, we trained 100 epochs, with an early termination by monitoring the validation error.}

\subsection{Detailed Results with 18-step input and output sequences}
\begin{table*}[ht]
\centering
\caption{Detailed experimental results with state-of-the-art models with 18-step input and prediction sequences.}
\resizebox{\linewidth}{!}{\begin{tabular}{c|c|c|c|c|c|c|c|c|c|c|c|}
\hline
Dataset                & T                       & Metric & HA         & MLP & STGCN & GCRNN     & DCRNN & ASTGCN & GWNet & GMAN & \toolname \\
\hline\hline 
 \multirow{18}{*}{NAVER-Seoul}    & \multirow{3}{*}{15min}  & MAE    &  6.54 &  5.28 &  4.63 &  4.87 &  4.86 &  5.09 &   4.91          &  5.20 &  \textbf{4.57}\\
                                  &                         & MAPE   & 18.24 & 16.86 & 14.49 & 15.23 & 15.35 & 16.14 &  14.86          & 16.98 & \textbf{14.43}\\
                                  &                         & RMSE   &  9.32 &  7.78 &  6.92 &  7.18 &  7.12 &  7.44 &   7.24          &  8.32 &  \textbf{6.72}\\
                                  \cline{2-12}
                                  & \multirow{3}{*}{30min}  & MAE    &  7.16 &  6.13 &  5.50 &  5.73 &  5.67 &  5.71 &   5.26          &  5.35 &  \textbf{5.04}\\
                                  &                         & MAPE   & 20.15 & 20.05 & 17.37 & 18.17 & 18.38 & 18.78 &  \textbf{16.16} & 17.47 & 16.34\\
                                  &                         & RMSE   & 10.18 &  9.51 &  8.83 &  9.03 &  8.80 &  8.73 &   8.13          &  8.67 &  \textbf{7.86}\\
                                  \cline{2-12}
                                  & \multirow{3}{*}{45min} & MAE    &  7.70 &  6.68 &  6.16 &  6.24 &  6.12 &  6.01 &  5.43           &  5.43 &  \textbf{5.18}\\
                                  &                        & MAPE   & 21.81 & 22.01 & 19.15 & 19.85 & 20.06 & 19.73 & \textbf{16.70}  & 17.72 & 16.81\\
                                  &                        & RMSE   & 10.89 & 10.48 &  9.95 &  9.99 &  9.61 &  9.28 &  8.53           &  8.84 &  \textbf{8.24}\\
                                  \cline{2-12}
                                  & \multirow{3}{*}{60min}  & MAE    &  8.22 &  7.08 &  6.77 &  6.58 &  6.40 &  6.22 &   5.55          &  5.48 &  \textbf{5.24}\\
                                  &                         & MAPE   & 23.37 & 23.44 & 20.42 & 20.95 & 21.09 & 20.37 &  16.97          & 17.89 & \textbf{16.94}\\
                                  &                         & RMSE   & 11.54 & 11.13 & 10.89 & 10.58 & 10.06 &  9.58 &   8.77          &  8.94 &  \textbf{8.39}\\
                                  \cline{2-12}
                                  & \multirow{3}{*}{75min} & MAE    &  8.73 &  7.44 &  7.40 &  6.87 &  6.63 &  6.46 &  5.68           &  5.53 &  \textbf{5.30}\\
                                  &                        & MAPE   & 24.91 & 24.76 & 21.62 & 21.91 & 21.93 & 20.93 & 17.31           & 18.05 & \textbf{17.10}\\
                                  &                        & RMSE   & 12.17 & 11.67 & 11.85 & 11.02 & 10.39 &  9.91 &  9.01           &  9.03 &  \textbf{8.50}\\
                                  \cline{2-12}
                                  & \multirow{3}{*}{90min}  & MAE    &  9.24 &  7.79 &  8.06 &  7.14 &  6.86 &  6.76 &   5.87          &  5.58 &  \textbf{5.40}\\
                                  &                         & MAPE   & 26.40 & 26.08 & 22.93 & 22.86 & 22.74 & 21.83 &  17.89          & 18.18 & \textbf{17.44}\\
                                  &                         & RMSE   & 12.77 & 12.17 & 12.86 & 11.43 & 10.69 & 10.32 &   9.33          &  9.09 &  \textbf{8.68}\\
                                  \cline{1-12}

\multirow{18}{*}{METR-LA}    & \multirow{3}{*}{15min}   & MAE    &  4.23 &  2.93 &  2.61 &  2.59 &  \textbf{2.56} &  3.25 &  2.72 &  2.86 &  2.66 \\
                             &                         & MAPE    &  9.76 &  7.76 &  \textbf{6.59} &  6.73 &  6.67 &  9.27 &  7.14 &  7.67 &  7.06 \\
                             &                         & RMSE    &  7.46 &  5.81 &  5.19 &  \textbf{5.12} &  5.10 &  6.28 &  5.20 &  5.77 &  5.28 \\
                             \cline{2-12}
                             & \multirow{3}{*}{30min}  & MAE     &  4.80 &  3.60 &  3.22 &  3.08 &  \textbf{3.01} &  3.80 &  3.12 &  3.14 &  3.02 \\
                             &                         & MAPE    & 11.30 & 10.00 &  \textbf{8.39} &  8.72 &  8.42 & 11.28 &  8.66 &  8.79 &  8.49 \\
                             &                         & RMSE    &  8.34 &  7.29 &  6.63 &  6.32 &  6.29 &  7.59 &  6.34 &  6.54 &  \textbf{6.28} \\
                             \cline{2-12}
                             & \multirow{3}{*}{45min}  & MAE    &  5.32 &  4.17 &  3.78 &  3.46 &  3.34 &  4.20 &  3.39 &  3.34 &  \textbf{3.24} \\
                             &                         & MAPE   & 12.71 & 11.93 &  9.84 & 10.26 &  9.71 & 12.64 &  9.63 &  9.55 &  \textbf{9.33} \\
                             &                         & RMSE   &  9.18 &  8.37 &  7.76 &  7.12 &  7.08 &  8.40 &  7.06 &  7.00 &  \textbf{6.87} \\
                             \cline{2-12}
                             & \multirow{3}{*}{60min}   & MAE    &  5.80 &  4.69 &  4.31 &  3.74 &  3.60 &  4.49 &  3.58 &  3.48 &  \textbf{3.40} \\
                             &                          & MAPE   & 14.04 & 13.68 & 11.13 & 11.50 & 10.73 & 13.69 & 10.30 & 10.10 &  \textbf{9.88} \\
                             &                          & RMSE   &  9.86 &  9.24 &  8.71 &  7.71 &  7.65 &  8.94 &  7.53 &  7.30 &  \textbf{7.24} \\
                             \cline{2-12}
                             & \multirow{3}{*}{75min}  & MAE    &  6.25 &  5.17 &  4.86 &  4.01 &  3.84 &  4.73 &  3.73 &  3.60 &  \textbf{3.53} \\
                             &                         & MAPE   & 15.26 & 15.47 & 12.45 & 12.59 & 11.68 & 14.60 & 10.90 & 10.56 & \textbf{10.31} \\
                             &                         & RMSE   & 10.46 &  9.95 &  9.62 &  8.21 &  8.15 &  9.34 &  7.87 &  7.52 &  \textbf{7.51} \\
                             \cline{2-12}
                             & \multirow{3}{*}{90min}   & MAE    &  6.65 &  5.58 &  5.41 &  4.23 & 4.06  &  4.97 &  3.85 &  3.71 &  \textbf{3.64} \\
                             &                          & MAPE   & 16.37 & 17.08 & 13.76 & 13.49 &12.53  & 15.53 & 11.39 & 11.00 & \textbf{10.74} \\
                             &                          & RMSE   & 10.97 & 10.52 & 10.47 &  8.79 & 8.58  &  9.71 &  8.12 &  7.71 &  \textbf{7.74} \\
                             
\hline
\end{tabular}}
\label{table:detailed_exp}
\end{table*}
\begin{figure}[ht]
    \centering
    \includegraphics[width=1\columnwidth]{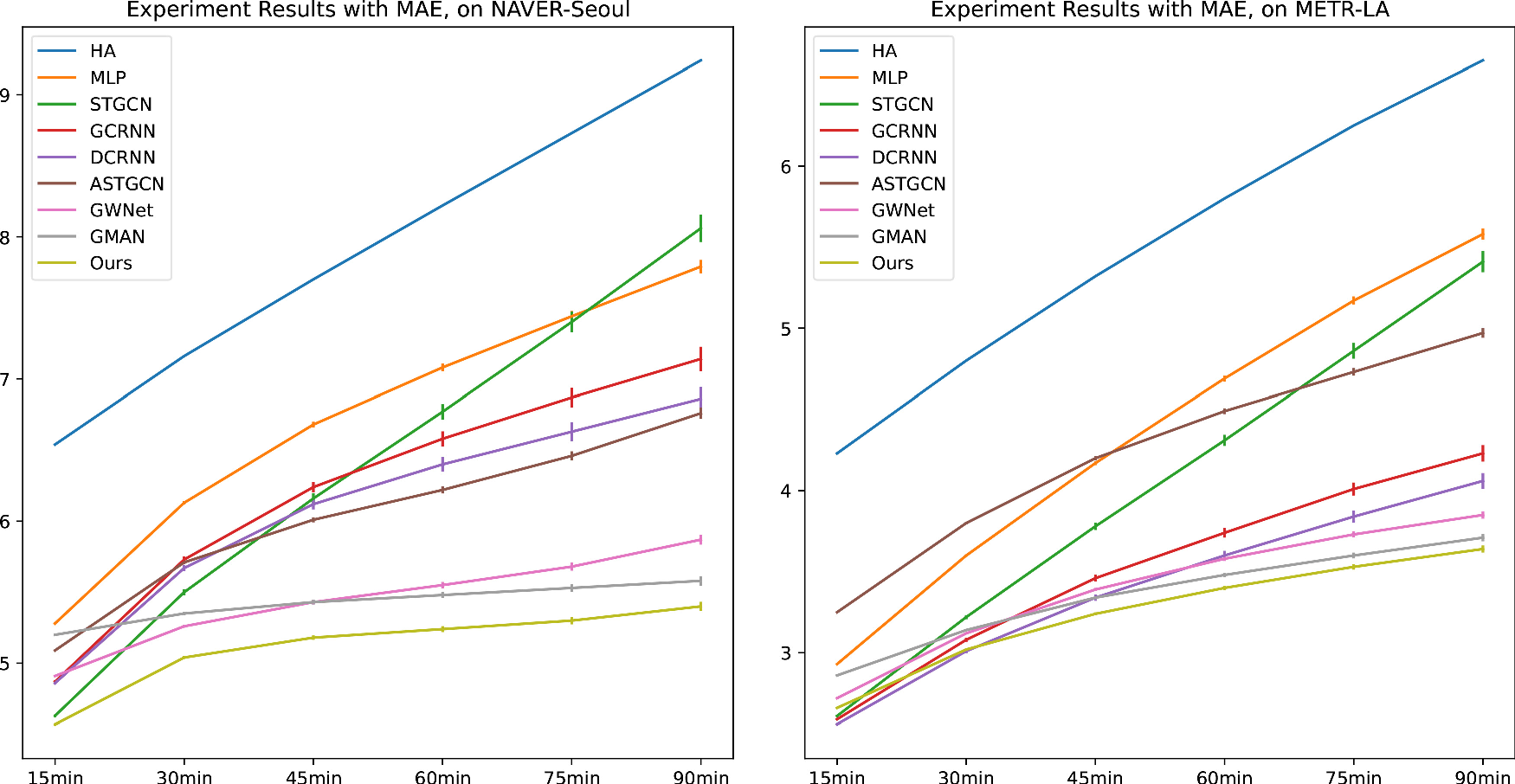}
    \caption{Performance result plots with error bars with NAVER-Seoul (left) and METR-LA (right) datasets.}
    \label{fig:errorbar}
\end{figure}
\begin{table}[htbp]
    \centering
    \caption{The computation times for each model with METR-LA}
    \resizebox{\linewidth}{!}{\begin{tabular}{c|c|c|c|c|c|c|c|c}
    \hline
    Computation Time     & DCRNN & GWNet & GMAN & \toolname & SimpleMem & CNN Decoder & RNN Decoder & \toolname w/ L = 1 \\
    \hline
    Training time per epoch (sec) & 691.32 & 102.06 & 500.19 & 196.6 & 169.6 & 104.78 & 39.94 & 127.53\\
    Inference Time (sec)          & 56.30 & 6.00 & 9.34 & 14.9 &12.14 & 10.2 & 4.26 & 10.77\\
    \hline
    \end{tabular}}
    \label{table:time}
\end{table}
\newpage

\subsection{Detailed Ablation Study Results}
\label{section:ablation_appendix}
\begin{table*}[ht]
    \centering
    \caption{Detailed ablation study result.}
\resizebox{\linewidth}{!}{\begin{tabular}{c|c|c|c|c|c|c|c|c|c|}
\hline
Dataset                & T                       & Metric & Ours & SimpleMem & CNN Decoder & RNN Decoder & Ours (L=1)  & Ours ($|\mathbb{P}| = k$) & Ours ($|\mathbb{P}| >> 1000$)\\
\hline\hline 
\multirow{18}{*}{NAVER-Seoul}     & \multirow{3}{*}{15min}  & MAE    &  4.57 &  5.72 &  \textbf{4.56} &  4.67 &  4.72           &  4.66 &   4.59\\
                                  &                         & MAPE   & 14.43 & 18.18 & \textbf{14.40} & 14.84 & 14.98           & 14.77 &  14.48\\
                                  &                         & RMSE   &  6.72 &  8.79 &  \textbf{6.71} &  6.83 &  6.87           &  6.89 &   6.73\\
                                  \cline{2-10}
                                  & \multirow{3}{*}{30min}  & MAE    &  \textbf{5.04} &  5.84 &  5.06 &  5.19 &  5.22           &  5.21 &   5.09\\
                                  &                         & MAPE   & \textbf{16.34} & 18.86 & 16.36 & 16.87 & 16.97           & 16.91 &  16.41\\
                                  &                         & RMSE   &  \textbf{7.86} &  9.24 &  7.90 &  8.02 &  8.03           &  8.18 &   7.97\\
                                  \cline{2-10}
                                  & \multirow{3}{*}{45min} & MAE    &  \textbf{5.18} &  6.10 &  5.23 &  5.38 &  5.43           &  5.43 &   5.27\\
                                  &                        & MAPE   & \textbf{16.81} & 19.98 & 16.98 & 17.61 & 17.66           & 17.72 &  16.97\\
                                  &                        & RMSE   &  \textbf{8.24} &  9.71 &  8.32 &  8.47 &  8.49           &  8.69 &   8.45\\
                                  \cline{2-10}
                                  & \multirow{3}{*}{60min}  & MAE    &  \textbf{5.24} &  6.38 &  5.32 &  5.47 &  5.52           &  5.53 &   5.36\\
                                  &                         & MAPE   & \textbf{16.94} & 21.42 & 17.19 & 17.91 & 17.97           & 18.05 &  17.19\\
                                  &                         & RMSE   &  \textbf{8.39} & 10.08 &  8.51 &  8.69 &  8.70           &  8.92 &   8.67\\
                                  \cline{2-10}
                                  & \multirow{3}{*}{75min} & MAE    &  \textbf{5.30} &  6.65 &  5.39 &  5.56 &  5.60           &  5.62 &   5.42\\
                                  &                        & MAPE   & \textbf{17.10} & 22.52 & 17.38 & 18.20 & 18.23           & 18.32 &  17.40\\
                                  &                        & RMSE   &  \textbf{8.50} & 10.48 &  8.62 &  8.86 &  8.85           &  9.09 &   8.80\\
                                  \cline{2-10}
                                  & \multirow{3}{*}{90min}  & MAE    &  \textbf{5.40} &  6.95 &  5.55 &  5.70 &  5.72           &  5.74 &   5.52\\
                                  &                         & MAPE   & \textbf{17.44} & 23.89 & 17.99 & 18.73 & 18.63           & 18.73 &  17.76\\
                                  &                         & RMSE   &  \textbf{8.68} & 10.88 &  8.82 &  9.10 &  9.05           &  9.31 &   8.94\\
                                  \cline{2-10}
\cline{1-1}
 \multirow{18}{*}{METR-LA}        & \multirow{3}{*}{15min}  & MAE    &  2.66 &  3.01 &  \textbf{2.63} &  2.68 &  2.68               &  2.67 &  2.68\\ 
                                  &                         & MAPE   &  7.06 &  8.03 &  \textbf{6.98} &  7.10 &  7.11               &  7.09 &  7.13\\ 
                                  &                         & RMSE   &  \textbf{5.28} &  5.94 &  5.32 &  5.31 &  5.31               &  5.35 &  5.31\\ 
                                  \cline{2-10}
                                  & \multirow{3}{*}{30min}  & MAE    &  3.02 &  3.27 &  \textbf{3.01} &  3.06 &  3.06               &  3.06 &  3.04\\ 
                                  &                         & MAPE   &  8.49 &  9.20 &  \textbf{8.46} &  8.56 &  8.59               &  8.67 &  8.51\\ 
                                  &                         & RMSE   &  6.28 &  6.68 &  6.36 &  6.32 &  \textbf{6.27}               &  6.36 &  6.32\\ 
                                  \cline{2-10}
                                  & \multirow{3}{*}{45min}  & MAE    &  \textbf{3.24} &  3.52 &  3.25 &  3.30 &  3.30              &  3.32 &  3.28\\ 
                                  &                         & MAPE   &  \textbf{9.33} & 10.18 &  9.30 &  9.44 &  9.47              &  9.67 &  9.31\\ 
                                  &                         & RMSE   &  6.87 &  7.28 &  6.94 &  6.93 &  \textbf{6.86}              &  7.00 &  6.93\\ 
                                  \cline{2-10}
                                  & \multirow{3}{*}{60min}  & MAE    &  \textbf{3.40} &  3.72 &  3.41 &  3.46 &  3.47               &  3.49 &  3.45\\ 
                                  &                         & MAPE   &  \textbf{9.88} & 10.94 &  \textbf{9.88} & 10.02 & 10.07      & 10.34 &  9.87\\
                                  &                         & RMSE   &  \textbf{7.24} &  7.70 &  7.28 &  7.31 & 7.25                &  7.39 &  7.32\\ 
                                  \cline{2-10}
                                  & \multirow{3}{*}{75min}  & MAE    &  3.53 &  3.91 &  \textbf{3.52} &  3.59 &  3.60              &  3.62 &  3.58\\ 
                                  &                         & MAPE   & \textbf{10.31} & 11.61 & 10.35 & 10.47 & 10.56              & 10.85 & 10.27\\
                                  &                         & RMSE   &  \textbf{7.51} &  8.06 &  7.50 &  7.59 &  7.53              &  7.67 &  7.59\\ 
                                 \cline{2-10}
                                  & \multirow{3}{*}{90min}  & MAE    &  \textbf{3.64} &  4.09 &  3.65 &  3.71 &  3.73               &  3.75 &  3.69\\ 
                                  &                         & MAPE   & \textbf{10.74} & 12.25 & 10.85 & 10.87 & 10.98               & 11.30 & 10.63\\
                                  &                         & RMSE   &  7.74 &  8.38 &  \textbf{7.71} &  7.81 &  7.75               &  7.91 &  7.81\\ 
                                  \cline{2-10}

\hline
\end{tabular}}
\label{table:detailed_ablation}
\end{table*}

\newpage
\begin{figure}[t]
    \centering
    \includegraphics[width=1\columnwidth]{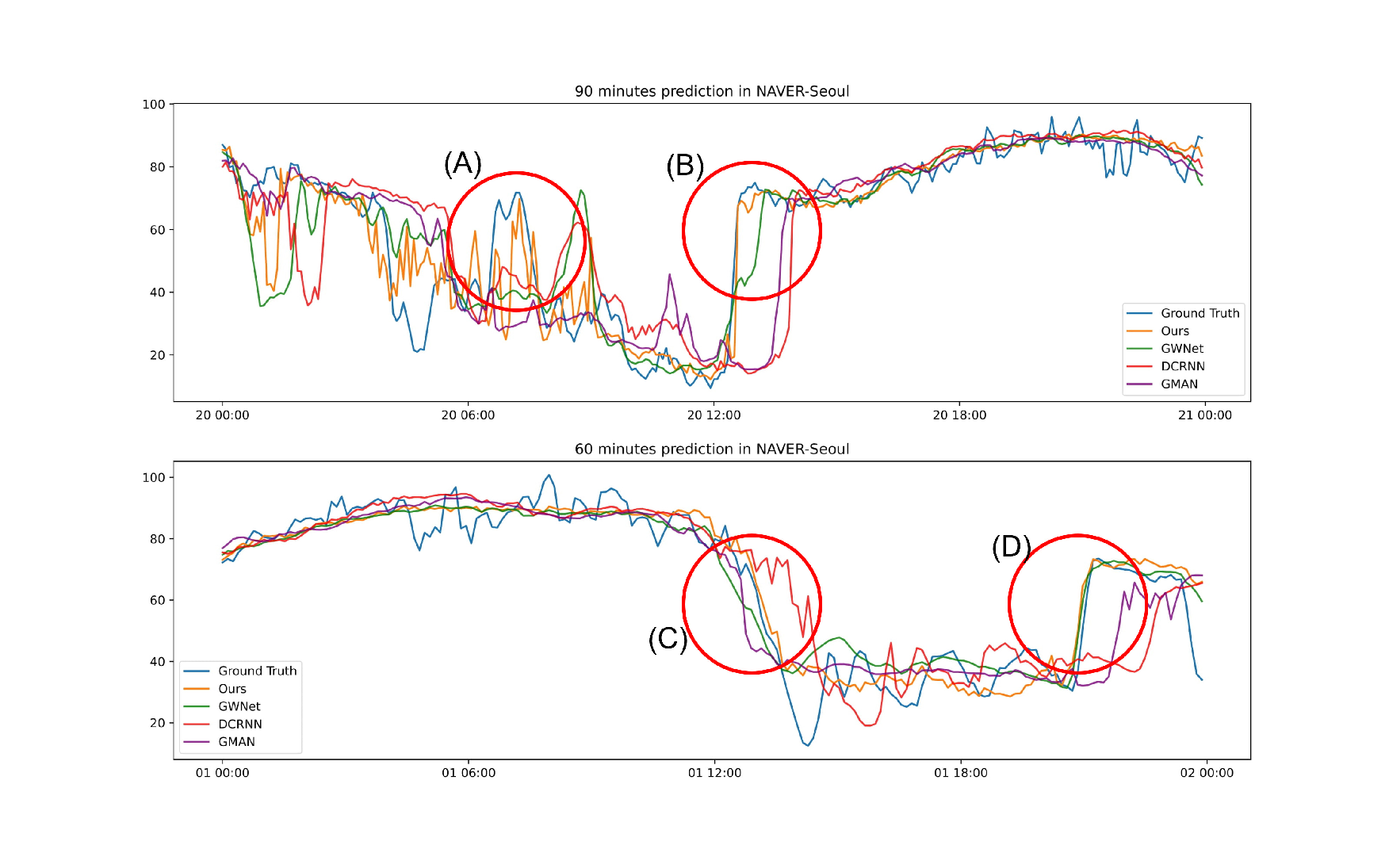}
    \caption{\rev{NAVER-Seoul speed prediction visualization for (upper) 90-minute forecasting and (lower) 60-minute forecasting. Red circles show the interval with abrupt speed changing, which are successfully detected by \toolname.}}
    \label{fig:qualitative}
\end{figure}

\subsection{Qualitative Evaluation}
\label{section:qualitative}
We present a qualitative analysis using the NAVER-Seoul dataset, which contains a complex road network and the dynamic traffic conditions of urban areas. 
In this analysis, we evaluate whether our approach effectively patterned traffic data for the traffic forecasting problem. 
If the approach is valid, we expect \toolname to be more accurate in predicting difficult trailing patterns due to the high correlation between leading and trailing patterns.
To verify our expectations, we visualize the long-term traffic prediction results, as shown in Figure~\ref{fig:qualitative} (top), where \toolname predicts the end of congestion at 12:00 PM (B) more accurately. Also, in 7:00 AM (A), \toolname quickly catches up on the unexpected peak speed (this continues for about 30 minutes, and \toolname efficiently catches up within 15 minutes). 
From Figure~\ref{fig:qualitative} (bottom), we can see that \toolname predicts a speed drop and its recovery more accurately compared to the other models. 
For example, GMAN and GWNet predict the speed drop earlier than the actual speed drop, and DCRNN predicts the speed drop later than the real speed drop, as shown in ~\ref{fig:qualitative} (C). 
However, \toolname predicts the occurrence of slowdowns on time, even for long-term traffic prediction (D). 
Overall, with representative traffic patterns and memorization, \toolname effectively handles abruptly changing traffic conditions, even for long-term predictions.

\begin{figure}[t]
    \centering
    \includegraphics[width=1\columnwidth]{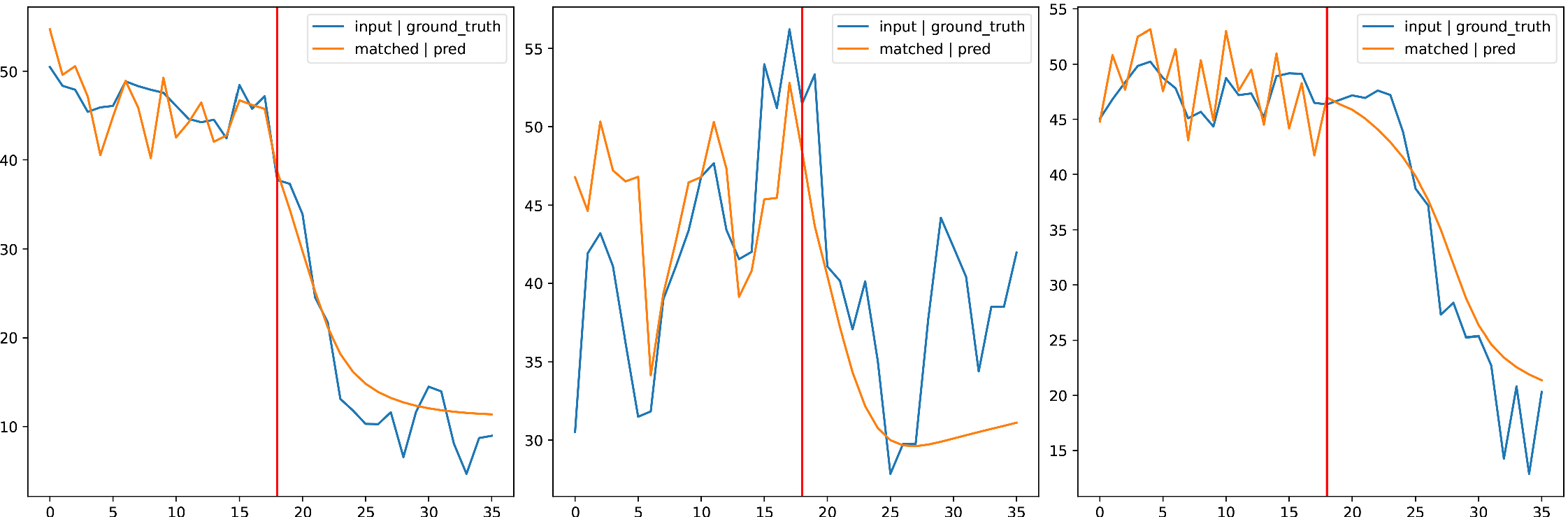}
    \caption{Sample input (ground truth, blue) and matched representative pattern for prediction (orange). the red line means the current time step $t$.}
    \label{fig:sample_inout}
\end{figure}

\rev{Next, we analyze how \toolname effectively matches input patterns with memorized patterns. 
Figure~\ref{fig:sample_inout} presents three examples, each of which has two time-series lines. 
The red line denotes the current time step $t$.
Before the current time step $t$, The blue and orange line means input and corresponding matched pattern, respectively.
After the current time step $t$, The blue and orange line means ground truth and prediction results, respectively.
In each example, we find that the model effectively retrieves a memorized pattern that matched the input sequence well for prediction. Note that all examples show how \toolname operates for difficult roads and time with rapid speed drops. 
}

\newpage
\subsection{Experiment with 12 Sequence Setting}
\begin{table*}[ht]
\centering
\caption{Experimental results with state-of-the-art models on common 12 sequence prediction setting. For PEMS-BAY, $\delta = 0.9$ and $|\mathbb{P}| \approx 100$.}
\resizebox{\linewidth}{!}{\begin{tabular}{c|c|c|c|c|c|c|c|c|c|c|c|}
\hline
Dataset                & T                       & Metric & HA         & MLP & STGCN & GCRNN     & DCRNN & ASTGCN & GWNet & GMAN & \toolname \\
\hline\hline 
 \multirow{12}{*}{NAVER-Seoul}    & \multirow{3}{*}{15min}  & MAE    &  6.22 &  5.28 &  4.69 &  5.00 &  4.92 &  4.91 &  \textbf{4.50} &  4.90 &  4.52\\
                                  &                         & MAPE   & 19.68 & 16.69 & 14.54 & 15.75 & 15.54 & 15.55 & 14.42 & 15.72 & \textbf{14.27}\\
                                  &                         & RMSE   &  9.54 &  7.78 &  7.02 &  7.34 &  7.20 &  7.23 &  \textbf{6.61} &  7.64 &  6.67\\
                                  \cline{2-12}
                                  & \multirow{3}{*}{30min}  & MAE    &  6.86 &  6.14 &  5.60 &  5.86 &  5.76 &  5.37 &  5.05 &  5.14 &  \textbf{5.01}\\
                                  &                         & MAPE   & 21.77 & 19.92 & 17.61 & 18.86 & 18.59 & 17.14 & 16.71 & 16.61 & \textbf{16.10}\\
                                  &                         & RMSE   & 10.66 &  9.53 &  8.99 &  9.15 &  8.92 &  8.31 &  \textbf{7.82} &  8.24 &  \textbf{7.82}\\
                                  \cline{2-12}
                                  & \multirow{3}{*}{60min}  & MAE    &  7.90 &  7.10 &  6.89 &  6.75 &  6.55 &  5.86 &  5.44 &  5.38 &  \textbf{5.30}\\
                                  &                         & MAPE   & 25.05 & 23.44 & 20.77 & 21.97 & 21.53 & 18.42 & 17.84 & 17.51 & \textbf{17.03}\\
                                  &                         & RMSE   & 12.28 & 11.16 & 11.01 & 10.75 & 10.29 &  9.16 &  8.57 &  8.71 &  \textbf{8.51}\\
                                  \cline{2-12}
                                  & \multirow{3}{*}{Avg.}   & MAE    &  6.87 &  5.82 &  5.39 &  5.55 &  5.43 &  5.16 &  4.74 &  5.10 &  \textbf{4.69}\\
                                  &                         & MAPE   & 21.77 & 18.84 & 16.67 & 17.76 & 17.49 & 16.79 & 15.44 & 16.48 & \textbf{14.97}\\
                                  &                         & RMSE   & 10.70 &  8.96 &  8.52 &  8.60 &  8.34 &  8.21 &  \textbf{7.28} &  8.12 &  \textbf{7.28}\\
                                  \cline{1-12}

\multirow{12}{*}{METR-LA}   & \multirow{3}{*}{15min}    & MAE    &  3.75 &  2.92 &  2.88 &  2.80 &  2.73 &  3.07 &  2.69 &  2.81        &   \textbf{2.65} \\
                             &                          & MAPE   & 10.01 &  7.71 &  7.62 &  7.50 &  7.12 &  5.90 &  \textbf{6.98} &  7.43&   7.01 \\
                             &                          & RMSE   &  7.31 &  5.82 &  5.74 &  5.51 &  5.27 &  8.23 &  \textbf{5.16} &  5.55&   5.29 \\
                             \cline{2-12}
                             & \multirow{3}{*}{30min}   & MAE    &  4.37 &  3.60 &  3.47 &  3.24 &  3.13 &  3.61 &  3.08 &  3.12         &   \textbf{3.03} \\
                             &                          & MAPE   & 11.85 &  9.93 &  9.57 &  9.00 &  8.65 & 10.34 &  8.43 &  \textbf{8.35}&   8.42 \\
                             &                          & RMSE   &  8.52 &  7.31 &  7.24 &  6.74 &  6.40 &  7.16 &  \textbf{6.21} &  6.46&   6.29 \\
                             \cline{2-12}
                             & \multirow{3}{*}{60min}   & MAE    &  5.45 &  4.70 &  4.59 &  3.81 &  3.58 &  4.42 &  3.53 &  \textbf{3.46}&   \textbf{3.46} \\
                             &                          & MAPE   & 15.04 & 13.64 & 12.70 & 10.90 & 10.43 & 13.35 & 10.05 & 10.06         &   \textbf{9.97} \\
                             &                          & RMSE   & 10.39 &  9.26 &  9.40 &  8.16 &  7.60 &  8.73 &  7.31 &  7.37         &   \textbf{7.29} \\
                              \cline{2-12}
                              & \multirow{3}{*}{Avg.}  & MAE    &   4.43 &  3.63 & 3.64 & 3.28 & 3.14 &  3.61 & 3.09 & 3.13 & \textbf{2.99}\\
                              &                        & MAPE   &  12.02 & 10.07 & 9.96 & 9.13 & 8.72 & 10.32 & 8.42 & 8.61 & \textbf{8.27}\\
                              &                        & RMSE   &   8.66 &  7.23 & 7.46 & 6.80 & 6.42 &  7.18 & 6.26 & 6.46 & \textbf{6.14}\\
                              \cline{1-12}
\multirow{12}{*}{PEMS-BAY}   & \multirow{3}{*}{15min}   & MAE   & 2.26 & 1.49 & 1.42& 1.34 & 1.33 & 1.55 & 1.30 & 1.36 & \textbf{1.27} \\
                             &                          & MAPE  & 5.03 & 3.15 & 3.10& 2.79 & 2.78 & 3.44 & \textbf{2.73} & 2.93 & 2.75 \\
                             &                          & RMSE  & 5.18 & 3.24 & 3.08& 2.82 & 2.79 & 3.17 & \textbf{2.74} & 2.88 & 2.80 \\
                             \cline{2-12}
                             & \multirow{3}{*}{30min}   & MAE   & 2.72 & 2.03 & 1.91& 1.72 & 1.68 & 2.01 & 1.63 & 1.64 & \textbf{1.62} \\
                             &                          & MAPE  & 6.18 & 4.55 & 4.49& 3.87 & 3.78 & 4.66 & 3.67 & 3.71 & \textbf{3.66} \\
                             &                          & RMSE  & 6.26 & 4.70 & 4.44& 3.85 & 3.77 & 4.19 & 3.70 & 3.78 & \textbf{3.65} \\
                             \cline{2-12}
                             & \multirow{3}{*}{60min}   & MAE   & 3.52 & 2.80 & 2.53& 2.08 & 2.01 & 2.57 & 1.95 & \textbf{1.90} & 1.95 \\
                             &                          & MAPE  & 8.19 & 6.86 & 6.23& 5.01 & 4.76 & 6.01 & 4.63 & \textbf{4.45} & 4.64 \\
                             &                          & RMSE  & 7.94 & 6.34 & 5.92& 4.72 & 4.59 & 5.27 & 4.52 & \textbf{4.40} & 4.51 \\
                              \cline{2-12}
                              & \multirow{3}{*}{Avg.}  & MAE    & 2.76 & 2.02 & 1.88& 1.66 & 1.62 & 1.97 & 1.57 & 1.58 & \textbf{1.55} \\
                              &                        & MAPE   & 6.29 & 4.61 & 4.28& 3.75 & 3.64 & 4.54 & 3.67 & \textbf{3.59} & 3.62 \\
                              &                        & RMSE   & 6.40 & 4.57 & 4.30& 3.66 & 3.58 & 4.18 & 3.47 & 3.58 & \textbf{3.44} \\
\hline
\end{tabular}}
\label{table:common_setting}
\end{table*}

\end{document}